\documentclass[11pt,a4paper]{article}
\usepackage[hyperref]{acl2017}
\usepackage{times}
\usepackage{latexsym}

\usepackage{url}

\usepackage{amsmath}
\usepackage{amssymb}
\usepackage{graphicx}
\newcommand{\cev}[1]{\reflectbox{\ensuremath{\vec{\reflectbox{\ensuremath{#1}}}}}}
\usepackage{framed}
\usepackage{subfigure}
\usepackage{booktabs}
\usepackage{empheq}
\usepackage{array}
\usepackage{color}
\usepackage{enumitem}
\newcommand{\unk}{$ \langle \textit{unk} \rangle $}

\newcommand{\ourModelName}{SEASS}

\aclfinalcopy 
\def\aclpaperid{***} 

\title{Selective Encoding for Abstractive Sentence Summarization}

\author{Qingyu Zhou$^\dag$\thanks{\; Contribution during internship at Microsoft Research.} \hspace{0.15cm} Nan Yang$^\ddag$ \hspace{0.15cm} Furu Wei$^\ddag$ \hspace{0.15cm}  Ming Zhou$^\ddag$ \\
	$^\dag$Harbin Institute of Technology, Harbin, China \\
	$^\ddag$Microsoft Research, Beijing, China \\
	{\tt qyzhou@hit.edu.cn} \hspace{0.25cm} {\tt \{nanya,fuwei,mingzhou\}@microsoft.com}
}

\date{}

\begin{document}
\maketitle
\begin{abstract}
	We propose a selective encoding model to extend the sequence-to-sequence framework for abstractive sentence summarization.
	It consists of a sentence encoder, a selective gate network, and an attention equipped decoder.
	The sentence encoder and decoder are built with recurrent neural networks.
	The selective gate network constructs a second level sentence representation by controlling the information flow from encoder to decoder.
	The second level representation is tailored for sentence summarization task, which leads to better performance.
	We evaluate our model on the English Gigaword, DUC 2004 and MSR abstractive sentence summarization datasets.
	The experimental results show that the proposed selective encoding model outperforms the state-of-the-art baseline models.
\end{abstract}

\section{Introduction}

Sentence summarization aims to shorten a given sentence and produce a brief summary of it.
This is different from document level summarization task since it is hard to apply existing techniques in extractive methods, such as extracting sentence level features and ranking sentences.
Early works propose using rule-based methods \citep{zajic2007multi}, syntactic tree pruning methods \citep{knight2002summarization}, statistical machine translation techniques \citep{banko2000headline} and so on for this task.
We focus on abstractive sentence summarization task in this paper.

Recently, neural network models have been applied in this task.
\citet{rush-chopra-weston:2015:EMNLP} use auto-constructed sentence-headline pairs to train a neural network summarization model.
They use a Convolutional Neural Network (CNN) encoder and feed-forward neural network language model decoder for this task.
\citet{chopra-auli-rush:2016:N16-1} extend their work by replacing the decoder with Recurrent Neural Network (RNN).
\citet{nallapatiabstractive} follow this line and change the encoder to RNN to make it a full RNN based sequence-to-sequence model \citep{sutskever2014sequence}.

\begin{figure}[ht]
	\centering
	\includegraphics[width=0.45\textwidth]{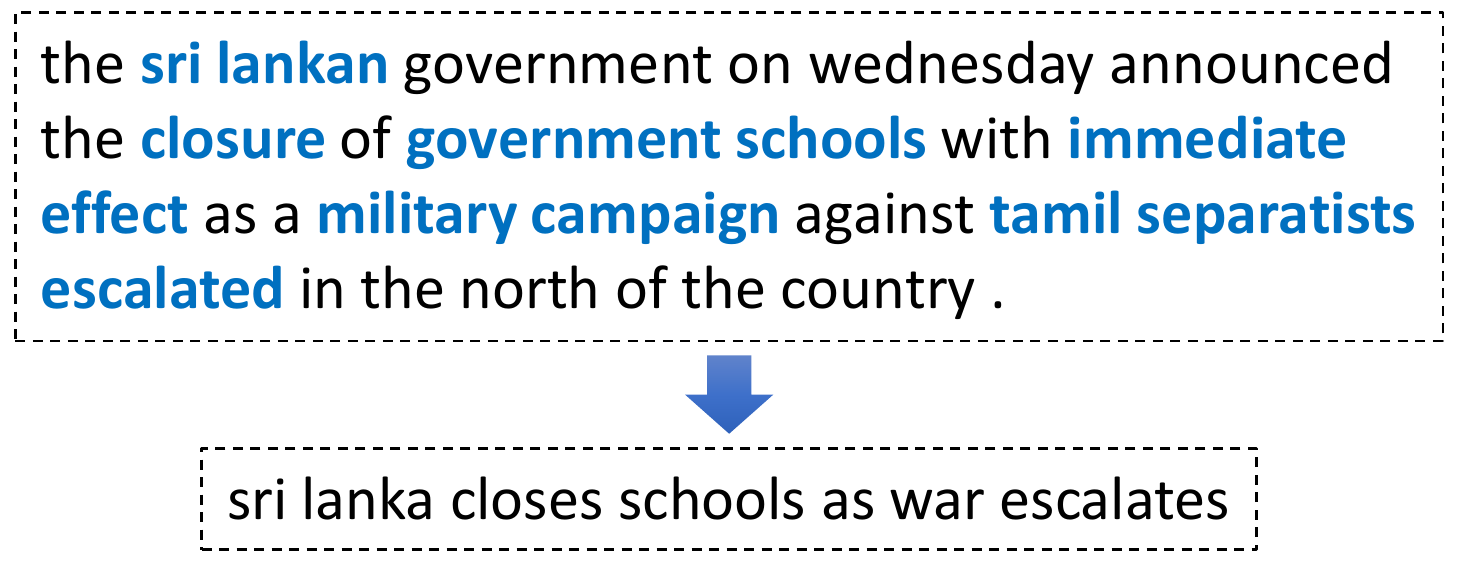}
	\caption{\label{fig:select}An abstractive sentence summarization system may produce the output summary by distilling the salient information from the highlight to generate a fluent sentence. We model the distilling process with selective encoding.
	}
\end{figure}

All the above works fall into the encoding-decoding paradigm, which first encodes the input sentence to an abstract representation and then decodes the intended output sentence based on the encoded information.
As an extension of the encoding-decoding framework, attention-based approach \citep{bahdanau2014neural} has been broadly used: the encoder produces a list of vectors for all tokens in the input, and the decoder uses an attention mechanism to dynamically extract encoded information and align with the output tokens.
This approach achieves huge success in tasks like machine translation, where alignment between all parts of the input and output are required.
However, in abstractive sentence summarization, there is no explicit alignment relationship between the input sentence and the summary except for the extracted common words.
The challenge here is not to infer the alignment, but to select the highlights while filtering out secondary information in the input.
A desired work-flow for abstractive sentence summarization is encoding, selection, and decoding.
After selecting the important information from an encoded sentence, the decoder produces the output summary using the selected information.
For example, in Figure \ref{fig:select}, given the input sentence, the summarization system first selects the important information, and then rephrases or paraphrases to produce a well-organized summary.
Although this is implicitly modeled in the encoding-decoding framework, we argue that abstractive sentence summarization shall benefit from explicitly modeling this selection process.

In this paper we propose \textbf{S}elective \textbf{E}ncoding for \textbf{A}bstractive \textbf{S}entence \textbf{S}ummarization (\textbf{SEASS}).
We treat the sentence summarization as a three-phase task: encoding, selection, and decoding.
It consists of a sentence encoder, a selective gate network, and a summary decoder.
First, the sentence encoder reads the input words through an RNN unit to construct the first level sentence representation.
Then the selective gate network selects the encoded information to construct the second level sentence representation.
The selective mechanism controls the information flow from encoder to decoder by applying a gate network according to the sentence information, which helps improve encoding effectiveness and release the burden of the decoder.
Finally, the attention-equipped decoder generates the summary using the second level sentence representation.
We conduct experiments on English Gigaword, DUC 2004 and Microsoft Research Abstractive Text Compression test sets.
Our \ourModelName{} model achieves 17.54 \textsc{Rouge}-2 F1, 9.56 \textsc{Rouge}-2 recall and 10.63 \textsc{Rouge}-2 F1 on these test sets respectively, which improves performance compared to the state-of-the-art methods.

\section{Related Work}
\label{sec:relWork}

Abstractive sentence summarization, also known as sentence compression and similar to headline generation, is used to help compress or fuse the selected sentences in extractive document summarization systems since they may inadvertently include unnecessary information.
The sentence summarization task has been long connected to the headline generation task.
There are some previous methods to solve this task, such as the linguistic rule-based method \citep{dorr2003hedge}.
As for the statistical machine learning based methods, \citet{banko2000headline} apply statistical machine translation techniques by modeling headline generation as a translation task and use 8000 article-headline pairs to train the system.

\citet{rush-chopra-weston:2015:EMNLP} propose leveraging news data in Annotated English Gigaword \citep{Napoles:2012:AG:2391200.2391218} corpus to construct large scale parallel data for sentence summarization task.
They propose an ABS model, which consists of an attentive Convolutional Neural Network encoder and an neural network language model \citep{bengio2003neural} decoder.
On this Gigaword test set and DUC 2004 test set, the ABS model produces the state-of-the-art results.
\citet{chopra-auli-rush:2016:N16-1} extend this work, which keeps the CNN encoder but replaces the decoder with recurrent neural networks.
Their experiments showes that the CNN encoder with RNN decoder model performs better than \citet{rush-chopra-weston:2015:EMNLP}.
\citet{nallapatiabstractive} further change the encoder to an RNN encoder, which leads to a full RNN sequence-to-sequence model.
Besides, they enrich the encoder with lexical and statistic features which play important roles in traditional feature based summarization systems, such as NER and POS tags, to improve performance.
Experiments on the Gigaword and DUC 2004 test sets show that the above models achieve state-of-the-art results.

\citet{gu-EtAl:2016:P16-1} and \citet{gulcehre-EtAl:2016:P16-1} come up similar ideas that summarization task can benefit from copying words from input sentences.
\citet{gu-EtAl:2016:P16-1} propose CopyNet to model the copying action in response generation, which also applies for summarization task.
\citet{gulcehre-EtAl:2016:P16-1} propose a switch gate to control whether to copy from source or generate from decoder vocabulary.
\citet{zeng2016efficient} also propose using copy mechanism and add a scalar weight on the gate of GRU/LSTM for this task.
\citet{cheng-lapata:2016:P16-1} use an RNN based encoder-decoder for extractive summarization of documents.

\citet{yu-buys-blunsom:2016:EMNLP2016} propose a segment to segment neural transduction model for sequence-to-sequence framework.
The model introduces a latent segmentation which determines correspondences between tokens of the input sequence and the output sequence.
Experiments on this task show that the proposed transduction model performs comparable to the ABS model.
\citet{shen-EtAl:2016:P16-1} propose to apply Minimum Risk Training (MRT) in neural machine translation to directly optimize the evaluation metrics.
\citet{DBLP:journals/corr/AyanaSLS16} apply MRT on abstractive sentence summarization task and the results show that optimizing for \textsc{Rouge} improves the test performance.

\section{Problem Formulation}
\label{sec:form}

For sentence summarization, given an input sentence $ x = (x_{1}, x_{2}, \dots, x_{n}) $, where $ n $ is the sentence length, $ x_{i} \in \mathcal{V}_{s}  $ and  $ \mathcal{V}_{s} $ is the source vocabulary, the system summarizes $ x $ by producing $ y = (y_{1}, y_{2}, \dots, y_{l}) $, where $ l \leq n $ is the summary length , $ y_{i} \in \mathcal{V}_{t} $ and $ \mathcal{V}_{t} $ is the target vocabulary.

If $ \vert y \vert \subseteq \vert x \vert $, which means all words in summary $ y $ must appear in given input, we denote this as \textit{extractive} sentence summarization.
If $ \vert y \vert \nsubseteq \vert x \vert $, which means not all words in summary come from input sentence, we denote this as \textit{abstractive} sentence summarization. Table \ref{probDefTable} provides an example.
We focus on \textit{abstracive} sentence summarization task in this paper.

\begin{table}[htbp]
	\begin{center}
		\small
		\setlength\extrarowheight{2pt}
		\begin{tabular}{|lp{0.35\textwidth}|}
			\hline
			\textbf{Input}: & South Korean President Kim Young-Sam left here Wednesday on a week - long state visit to Russia and Uzbekistan for talks on North Korea 's nuclear confrontation and ways to strengthen bilateral ties .\\
			\textbf{Output}: & Kim leaves for Russia for talks on NKorea nuclear standoff \\
			\hline
		\end{tabular}
	\end{center}
	\caption{\label{probDefTable} An abstractive sentence summarization example.}
\end{table}

\section{Model}
\label{sec:model}

\begin{figure*}[htb]
	\centering
	\includegraphics[scale=0.6]{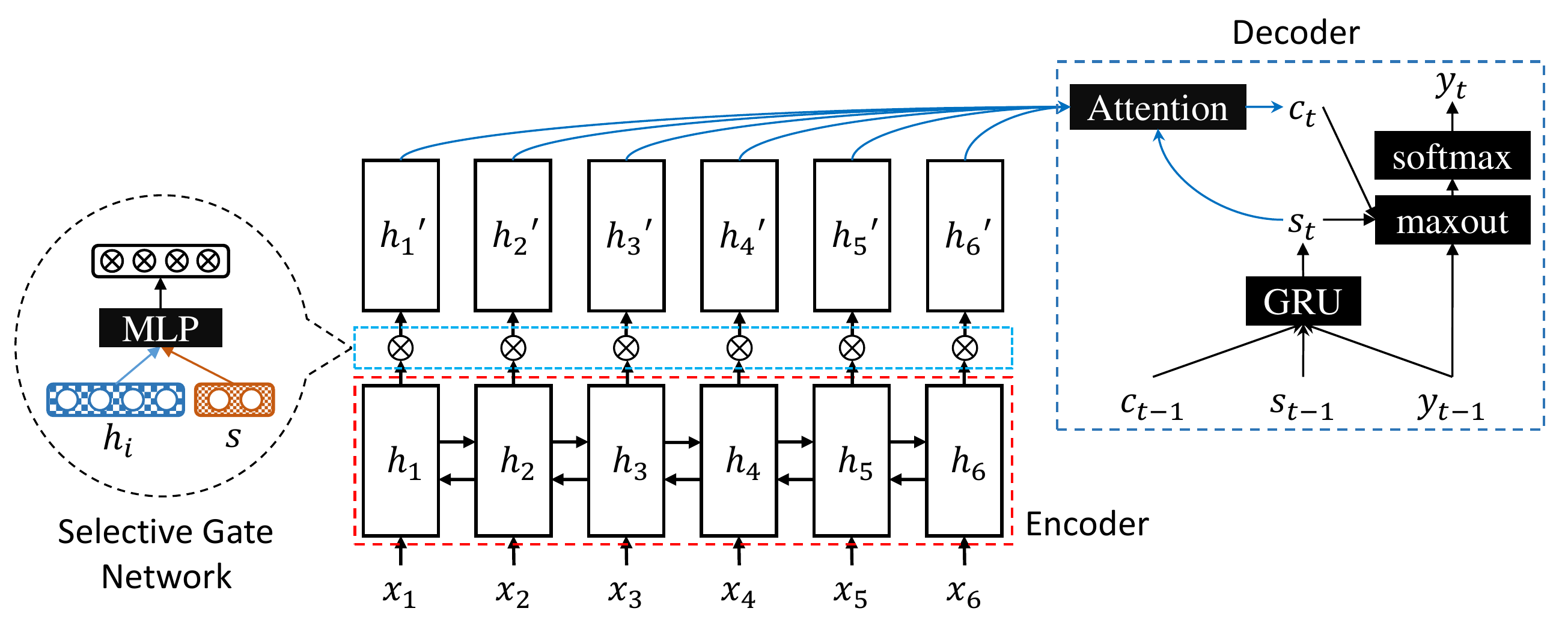}
	\caption{\label{fig:model} Overview of the \textbf{S}elective \textbf{E}ncoding for \textbf{A}bstractive \textbf{S}entence \textbf{S}ummarization (\textbf{SEASS}).}
\end{figure*}

As shown in Figure \ref{fig:model}, our model consists of a sentence encoder using the Gated Recurrent Unit (GRU) \cite{cho-EtAl:2014:EMNLP2014}, a selective gate network and an attention-equipped GRU decoder.
First, the bidirectional GRU encoder reads the input words $ x = (x_{1}, x_{2}, \dots, x_{n}) $ and builds its representation $ (h_{1}, h_{2}, \dots, h_{n}) $.
Then the selective gate selects and filters the word representations according to the sentence meaning representation to produce a tailored sentence word representation for abstractive sentence summarization task.
Lastly, the GRU decoder produces the output summary with attention to the tailored representation.
In the following sections, we introduce the sentence encoder, the selective mechanism, and the summary decoder respectively.

\subsection{Sentence Encoder}
The role of the sentence encoder is to read the input sentence and construct the basic sentence representation.
Here we employ a bidirectional GRU (BiGRU) as the recurrent unit, where GRU is defined as:
\begin{empheq}{align}
z_i &= \sigma(\textbf{W}_z[x_i,h_{i-1}]) \\
r_i &= \sigma(\textbf{W}_r[x_i,h_{i-1}]) \\
\widetilde{h}_i &= \tanh(\textbf{W}_h[x_i,r_i \odot h_{i-1}]) \\
h_i &= (1-z_i)\odot h_{i-1} + z_i \odot \widetilde{h}_i 
\end{empheq}
where $ \textbf{W}_z $, $ \textbf{W}_r $ and $ \textbf{W}_h $ are weight matrices.

The BiGRU consists of a forward GRU and a backward GRU.
The forward GRU reads the input sentence word embeddings from left to right and gets a sequence of hidden states, $ (\vec{h}_{1}, \vec{h}_{2}, \dots, \vec{h}_{n})  $.
The backward GRU reads the input sentence embeddings reversely, from right to left, and results in another sequence of hidden states, $ (\cev{h}_{1}, \cev{h}_{2}, \dots, \cev{h}_{n}) $:
\begin{empheq}{align}
\vec{h}_{i} =\text{GRU}(x_{i}, \vec{h}_{i-1})\\
\cev{h}_{i} = \text{GRU}(x_{i}, \cev{h}_{i+1})
\end{empheq}

The initial states of the BiGRU are set to zero vectors, i.e., $ \vec{h}_{1} = 0 $ and $ \cev{h}_{n} = 0 $.
After reading the sentence, the forward and backward hidden states are concatenated, i.e., $ h_{i} = [\vec{h}_{i} ; \cev{h}_{i}] $, to get the basic sentence representation.

\subsection{Selective Mechanism}
In the sequence-to-sequence machine translation (MT) model, the encoder and decoder are responsible for mapping input sentence information to a list of vectors and decoding the sentence representation vectors to generate an output sentence \cite{bahdanau2014neural}.
Some previous works apply this framework to summarization generation tasks \citep{nallapatiabstractive,gu-EtAl:2016:P16-1,gulcehre-EtAl:2016:P16-1}.
However, abstractive sentence summarization is different from MT in two ways.
First, there is no explicit alignment relationship between the input sentence and the output summary except for the common words.
Second, summarization task needs to keep the highlights and remove the unnecessary information, while MT needs to keep all information literally.

Herein, we propose a selective mechanism to model the selection process for abstractive sentence summarization.
The selective mechanism extends the sequence-to-sequence model by constructing a tailored representation for abstractive sentence summarization task.
Concretely, the selective gate network in our model takes two vector inputs, the sentence word vector $ h_{i} $ and the sentence representation vector $ s $.
The sentence word vector $ h_{i} $ is the output of the BiGRU encoder and represents the meaning and context information of word $ x_{i} $.
The sentence vector $ s $ is used to represent the meaning of the sentence.
For each word $ x_{i} $, the selective gate network generates a gate vector $ sGate_{i} $ using $ h_{i} $ and $ s $, then the tailored representation is constructed, i.e., $ h_{i}' $.

In detail, we concatenate the last forward hidden state $ \vec{h}_{n} $ and backward hidden state $ \cev{h}_{1} $ as the sentence representation $ s $:
\begin{equation}
s = \left[
\begin{matrix}
\cev{h}_{1}\\
\vec{h}_{n}
\end{matrix}
\right]
\end{equation}

For each time step $ i $, the selective gate takes the sentence representation $ s $ and BiGRU hidden $ h_{i} $ as inputs to compute the  gate vector $ sGate_{i} $:
\begin{empheq}{align}
sGate_{i} &= \sigma(\mathbf{W}_{s}h_{i} + \mathbf{U}_{s}s + b)\\
h_{i}' &= h_{i} \odot sGate_{i}
\end{empheq}
where $ \mathbf{W}_{s} $ and $ \mathbf{U}_{s} $ are weight matrices, $ b $ is the bias vector, $ \sigma $ denotes sigmoid activation function, and $ \odot $ is element-wise multiplication.
After the selective gate network, we obtain another sequence of vectors $ (h_{1}', h_{2}', \dots, h_{n}') $.
This new sequence is then used as the input sentence representation for the decoder to generate the summary.

\subsection{Summary Decoder}
On top of the sentence encoder and the selective gate network, we use GRU with attention as the decoder to produce the output summary.

At each decoding time step $ t $, the GRU reads the previous word embedding $ w_{t-1} $ and previous context vector $ c_{t -1} $ as inputs to compute the new hidden state $ s_{t} $. To initialize the GRU hidden state, we use a linear layer with the last backward encoder hidden state $ \cev{h}_{1} $ as input:
\begin{empheq}{align}
s_{t} &= \text{GRU}(w_{t-1}, c_{t - 1}, s_{t-1})\\
s_{0} &= \tanh (\mathbf{W}_{d}\cev{h}_{1} + b)
\end{empheq}
where $ \mathbf{W}_{d} $ is the weight matrix and $ b $ is the bias vector.

The context vector $ c_{t} $ for current time step $ t $ is computed through the concatenate attention mechanism \citep{luong-pham-manning:2015:EMNLP}, which matches the current decoder state $ s_{t} $ with each encoder hidden state $ h_{i}' $ to get an importance score. The importance scores are then normalized to get the current context vector by weighted sum:
\begin{empheq}{align}
e_{t,i} &= v_{a}^{\top}\tanh(\mathbf{W}_{a}s_{t-1} + \mathbf{U}_{a}h_{i}')\\
\alpha_{t,i} &= \frac{\exp (e_{t,i})}{\sum_{i=1}^{n}\exp (e_{t,i})}\\
c_{t} &= \sum_{i = 1}^{n} \alpha_{t,i}h_{i}'
\end{empheq}

We then combine the previous word embedding $ w_{t-1} $, the current context vector $ c_{t} $, and the decoder state $ s_{t} $ to construct the readout state $ r_{t} $.
The readout state is then passed through a maxout hidden layer \citep{goodfellow2013maxout} to predict the next word with a softmax layer over the decoder vocabulary.
\begin{empheq}{align}
r_{t} &= \mathbf{W}_{r}w_{t-1} + \mathbf{U}_{r}c_{t} + \mathbf{V}_{r}s_{t}\\
\label{eq:maxout}m_{t} &= [\max\{r_{t, 2j-1}, r_{t, 2j}\}]^{\top}_{j = 1,\dots, d}\\
p(y_{t} &\vert y_{1}, \dots, y_{t-1}) = \text{softmax}(\mathbf{W}_{o}m_{t})
\end{empheq}
where $ \mathbf{W}_{a} $, $ \mathbf{U}_{a} $, $ \mathbf{W}_{r} $, $ \mathbf{U}_{r} $, $ \mathbf{V}_{r} $ and $ \mathbf{W}_{o} $ are weight matrices.
Readout state $ r_{t} $ is a $ 2d $-dimensional vector, and the maxout layer (Equation \ref{eq:maxout}) picks the max value for every two numbers in $ r_{t} $ and produces a d-dimensional vector $ m_{t} $.

\subsection{Objective Function}
Our goal is to maximize the output summary probability given the input sentence.
Therefore, we optimize the negative log-likelihood loss function:
\begin{equation}
J(\theta) = - \frac{1}{\vert \mathcal{D} \vert} \sum_{(x, y) \in \mathcal{D}} \log p(y|x)
\end{equation}
where $ \mathcal{D} $ denotes a set of parallel sentence-summary pairs and $ \theta $ is the model parameter.
We use Stochastic Gradient Descent (SGD) with mini-batch to learn the model parameter $ \theta $.

\newcommand{\significant}{$ {}^{\text{\textbf{-}}} $}
\newcommand{\otherpaper}{$ {}^{\ddag} $}

\section{Experiments}
\label{sec:expr}
In this section we introduce the dataset we use, the evaluation metric, the implementation details, the baselines we compare to, and the performance of our system.
\subsection{Dataset}
\paragraph{Training Set}
For our training set, we use a parallel corpus which is constructed from the Annotated English Gigaword dataset \citep{Napoles:2012:AG:2391200.2391218} as mentioned in \citet{rush-chopra-weston:2015:EMNLP}.
The parallel corpus is produced by pairing the first sentence and the headline in the news article with some heuristic rules. 
We use the script\footnote{\label{ft:absCode}\url{https://github.com/facebook/NAMAS}} released by \citet{rush-chopra-weston:2015:EMNLP} to pre-process and extract the training and development datasets.
The script performs various basic text normalization, including PTB tokenization, lower-casing, replacing all digit characters with \#, and replacing word types seen less than 5 times with \unk{}.
The extracted corpus contains about 3.8M sentence-summary pairs for the training set and 189K examples for the development set.

For our test set, we use the English Gigaword, DUC 2004, and Microsoft Research Abstractive Text Compression test sets.

\paragraph{English Gigaword Test Set}

We randomly sample 8000 pairs from the extracted development set as our development set since it is relatively large.
For the test set, we use the same randomly held-out test set of 2000 sentence-summary pairs as \citet{rush-chopra-weston:2015:EMNLP}.\footnote{Thanks to \citet{rush-chopra-weston:2015:EMNLP}, we acquired the test set they used. Following \citet{chopra-auli-rush:2016:N16-1}, we remove pairs with empty titles resulting in slightly different accuracy compared to \citet{rush-chopra-weston:2015:EMNLP} for their systems. The cleaned test set contains 1951 sentence-summary pairs.}

We also find that except for the empty titles, this test set has some invalid lines like the input sentence containing only one word.
Therefore, we further sample 2000 pairs as our internal test set and release it for future works\footnote{Our development and test sets can be found at \url{https://res.qyzhou.me}}.
\paragraph{DUC 2004 Test Set}
We employ DUC 2004 data for tasks 1 \& 2 \citep{over2007duc} in our experiments as one of the test sets since it is too small to train a neural network model on.
The dataset pairs each document with 4 different human-written reference summaries which are capped at 75 bytes.
It has 500 input sentences with each sentence paired with 4 summaries.

\paragraph{MSR-ATC Test Set}
\citet{toutanova-EtAl:2016:EMNLP2016} release a new dataset for sentence summarization task by crowdsourcing.
This dataset contains approximately 6,000 source text sentences with multiple manually-created summaries (about 26,000 sentence-summary pairs in total).
\citet{toutanova-EtAl:2016:EMNLP2016} provide a standard split of the data into training, development, and test sets, with 4,936, 448 and 785 input sentences respectively.
Since the training set is too small, we only use the test set as one of our test sets.
We denote this dataset as MSR-ATC (Microsoft Research Abstractive Text Compression) test set in the following.

Table \ref{dataTable} summarizes the statistic information of the three datasets we used.

\begin{table}[htbp]
	\begin{center}
		\begin{tabular}{llll}
			\toprule
			\bf Data Set & Giga & DUC$ {}^{\dag} $ & MSR$ {}^{\dag} $ \\ 
			\midrule
			\#(sent) & 3.99M  & 500 & 785  \\
			\#(sentWord)  & 125M & 17.8K & 29K \\
			\#(summWord)  & 33M & 20.9K & 85.9K \\
			\#(ref) & 1 & 4 & 3-5  \\
			AvgInputLen & 31.35 & 35.56 & 36.97 \\
			AvgSummLen & 8.23 & 10.43 & 25.5 \\
			\bottomrule
		\end{tabular}
	\end{center}
	\caption{\label{dataTable} Data statistics for the English Gigaword, DUC 2004 and MSR-ATC datasets.
		\#($ x $) denotes the number of $ x $, e.g., \#(ref) is the number of reference summaries of an input sentence. AvgInputLen is the average input sentence length and AvgSummLen is the average summary length.
		\dag DUC 2004 and MSR-ATC datasets are for test purpose only.}
\end{table}

\subsection{Evaluation Metric}
We employ \textsc{Rouge} \citep{lin2004rouge} as our evaluation metric.
\textsc{Rouge} measures the quality of summary by computing overlapping lexical units, such as unigram, bigram, trigram, and longest common subsequence (LCS).
It becomes the standard evaluation metric for DUC shared tasks and popular for summarization evaluation.
Following previous work, we use \textsc{Rouge}-1 (unigram), \textsc{Rouge}-2 (bigram) and \textsc{Rouge}-L (LCS) as the evaluation metrics in the reported experimental results.

\subsection{Implementation Details}

\paragraph{Model Parameters}
The input and output vocabularies are collected from the training data, which have 119,504 and 68,883 word types respectively.
We set the word embedding size to 300 and all GRU hidden state sizes to 512.
We use dropout \citep{srivastava2014dropout} with probability $ p = 0.5 $.

\paragraph{Model Training}
We initialize model parameters randomly using a Gaussian distribution with Xavier scheme \citep{glorot2010understanding}.
We use Adam \citep{kingma2014adam} as our optimizing algorithm.
For the hyperparameters of Adam optimizer, we set the learning rate $ \alpha = 0.001 $, two momentum parameters $ \beta_{1} = 0.9 $ and $ \beta_{2} = 0.999 $ respectively, and $ \epsilon=10^{-8} $.
During training, we test the model performance (\textsc{Rouge}-2 F1) on development set for every 2,000 batches.
We halve the Adam learning rate $ \alpha $ if the \textsc{Rouge}-2 F1 score drops for twelve consecutive tests on development set.
We also apply gradient clipping \citep{pascanu2013difficulty} with range $ [-5, 5] $ during training.
To both speed up the training and converge quickly, we use mini-batch size 64 by grid search.

\paragraph{Beam Search}
We use beam search to generate multiple summary candidates to get better results.
To avoid favoring shorter outputs, we average the ranking score along the beam path by dividing it by the number of generated words.
To both decode fast and get better results, we set the beam size to 12 in our experiments.

\subsection{Baseline}

We compare \ourModelName{} model with the following state-of-the-art baselines:
\begin{description}[noitemsep]
	\item[ABS] \citet{rush-chopra-weston:2015:EMNLP} use an attentive CNN encoder and NNLM decoder to do the sentence summarization task.
	We trained this baseline model with the released code\textsuperscript{\ref{ft:absCode}} and evaluate it with our internal English Gigaword test set and MSR-ATC test set. 
	\item[ABS+] Based on ABS model, \citet{rush-chopra-weston:2015:EMNLP} further tune their model using DUC 2003 dataset, which leads to improvements on DUC 2004 test set.
	\item[CAs2s] As an extension of the ABS model, \citet{chopra-auli-rush:2016:N16-1} use a convolutional attention-based encoder and RNN decoder, which outperforms the ABS model.
	\item[Feats2s] \citet{nallapatiabstractive} use a full RNN sequence-to-sequence encoder-decoder model and add some features to enhance the encoder, such as POS tag, NER, and so on.
	\item[Luong-NMT] Neural machine translation model of \citet{luong-pham-manning:2015:EMNLP} with two-layer LSTMs for the encoder-decoder with 500 hidden units in each layer implemented in \cite{chopra-auli-rush:2016:N16-1}. 
	\item[s2s+att] We also implement a sequence-to-sequence model with attention as our baseline and denote it as ``s2s+att''.
\end{description}

\subsection{Results}
We report \textsc{Rouge} F1, \textsc{Rouge} recall and \textsc{Rouge} F1 for English Gigaword, DUC 2004 and MSR-ATC test sets respectively.
We use the official \textsc{Rouge} script (version 1.5.5) \footnote{http://www.berouge.com/} to evaluate the summarization quality in our experiments.
For English Gigaword\footnote{The \textsc{Rouge} evaluation option is the same as \citet{rush-chopra-weston:2015:EMNLP}, -m -n 2 -w 1.2} and MSR-ATC\footnote{The \textsc{Rouge} evaluation option is, -m -n 2 -w 1.2} test sets, the outputs have different lengths so we evaluate the system with F1 metric.
As for the DUC 2004 test set\footnote{The \textsc{Rouge} evaluation option is, -m -b 75 -n 2 -w 1.2}, the task requires the system to produce a fixed length summary (75 bytes), therefore we employ \textsc{Rouge} recall as the evaluation metric.
To satisfy the length requirement,  we decode the output summary to a roughly expected length following \citet{rush-chopra-weston:2015:EMNLP}.

\paragraph{English Gigaword}
We acquire the test set from \citet{rush-chopra-weston:2015:EMNLP} so we can make fair comparisons to the baselines.

\begin{table}[htbp]
	\begin{center}
		\begin{tabular}{llll}
			\toprule
			\bf Models & \bf RG-1 & \bf RG-2 & \bf RG-L \\ 
			\midrule
			ABS (beam)\otherpaper{} & 29.55\significant{} & 11.32\significant{}  & 26.42\significant{} \\
			ABS+ (beam)\otherpaper{} & 29.76\significant{} & 11.88\significant{} & 26.96\significant{} \\
			Feats2s (beam)\otherpaper{} & 32.67\significant{} & 15.59\significant{} & 30.64\significant{} \\
			CAs2s (greedy)\otherpaper{} & 33.10\significant{} & 14.45\significant{} & 30.25\significant{} \\
			CAs2s (beam)\otherpaper{} & 33.78\significant{} & 15.97\significant{} & 31.15\significant{} \\
			Luong-NMT (beam)\otherpaper{} & 33.10\significant{} & 14.45\significant{} & 30.71\significant{} \\
			s2s+att (greedy) & 33.18\significant{} & 14.79\significant{}  & 30.80\significant{} \\
			s2s+att (beam) & 34.04\significant{} & 15.95\significant{}  & 31.68\significant{} \\
			\hline
			\ourModelName{} (greedy) & 35.48  &  16.50  &  32.93 \\
			\ourModelName{} (beam) & \textbf{36.15} & \textbf{17.54} & \textbf{33.63} \\
			
			\bottomrule
		\end{tabular}
	\end{center}
	\caption{\label{gw_fb-table} Full length \textsc{Rouge} F1 evaluation results on the English Gigaword test set used by \citet{rush-chopra-weston:2015:EMNLP}. RG in the Table denotes \textsc{Rouge}. Results with \otherpaper{} mark are taken from the corresponding papers. The superscript \significant{}  indicates that our \ourModelName{} model with beam search performs significantly better than it as given by the 95\% confidence interval in the official \textsc{Rouge} script.}
\end{table}

\begin{table}[h]
	\begin{center}
		\begin{tabular}{lccc}
			\toprule
			\bf Models & \bf RG-1 & \bf RG-2 & \bf RG-L \\ 
			\midrule
			ABS (beam) & 37.41\significant{} & 15.87\significant{} & 34.70\significant{} \\
			s2s+att (greedy) & 42.41\significant{} & 20.76\significant{} & 39.84\significant{} \\
			s2s+att (beam) & 43.76\significant{} & 22.28\significant{} & 41.14\significant{} \\
			\hline
			\ourModelName{} (greedy) & 45.27 & 22.88 & 42.20\\
			\ourModelName{} (beam) & \textbf{46.86} & \textbf{24.58}  & \textbf{43.53} \\
			
			\bottomrule
		\end{tabular}
	\end{center}
	\caption{\label{gw-table} Full length \textsc{Rouge} F1 evaluation on our internal English Gigaword test data. The superscript \significant{}  indicates that our \ourModelName{} model performs significantly better than it as given by the 95\% confidence interval in the official \textsc{Rouge} script. }
\end{table}

In Table \ref{gw_fb-table}, we report the \textsc{Rouge} F1 score of our model and the baseline methods.
Our \ourModelName{} model with beam search outperforms all baseline models by a large margin.
Even for greedy search, our model still performs better than other methods which used beam search.
For the popular \textsc{Rouge}-2 metric, our \ourModelName{} model achieves 17.54 F1 score and performs better than the previous works.
Compared to the ABS model, our model has a 6.22 \textsc{Rouge}-2 F1 relative gain.
Compared to the highest CAs2s baseline, our model achieves 1.57 \textsc{Rouge}-2 F1 improvement and passes the significant test according to the official \textsc{Rouge} script.

Table \ref{gw-table} summarizes our results on our internal test set using \textsc{Rouge} F1 evaluation metrics.
The performance on our internal test set is comparable to our development set, which achieves 24.58 \textsc{Rouge}-2 F1 and outperforms the baselines.

\paragraph{DUC 2004}
We evaluate our model using the \textsc{Rouge} recall score since the reference summaries of the DUC 2004 test set are capped at 75 bytes.
Therefore, we decode the summary to a fixed length 18 to ensure that the generated summary satisfies the minimum length requirement.
As summarized in Table \ref{duc-table}, our \ourModelName{} outperforms all the baseline methods and achieves 29.21, 9.56 and 25.51 for \textsc{Rouge} 1, 2 and L recall.
Compared to the ABS+ model which is tuned using DUC 2003 data, our model performs significantly better by 1.07 \textsc{Rouge}-2 recall score and is trained only with English Gigaword sentence-summary data without being tuned using DUC data.

\begin{table}[htb]
	\begin{center}
		\setlength\tabcolsep{5pt}
		\begin{tabular}{lccc}
			\toprule
			\bf Models & \bf RG-1 & \bf RG-2 & \bf RG-L \\ 
			\midrule
			ABS (beam)\otherpaper{} & 26.55\significant{}  & 7.06\significant{}  & 22.05\significant{} \\
			ABS+ (beam)\otherpaper & 28.18\significant{}  & 8.49\significant{}  & 23.81\significant{} \\
			Feats2s (beam)\otherpaper & 28.35\significant{} & 9.46  & 24.59\significant{} \\
			CAs2s (greedy)\otherpaper & 29.13  & 7.62\significant{}  & 23.92\significant{} \\
			CAs2s (beam)\otherpaper & 28.97  & 8.26\significant{}  & 24.06\significant{} \\
			Luong-NMT (beam)\otherpaper{} & 28.55 & 8.79\significant{}  & 24.43\significant{} \\
			s2s+att (greedy) & 27.03\significant{} & 7.89\significant{} &  23.80\significant{} \\
			s2s+att (beam) & 28.13 & 9.25 &  24.76 \\
			\hline
			\ourModelName{} (greedy) & 28.68 & 8.55  & 25.04 \\
			\ourModelName{} (beam) & \textbf{29.21} & \textbf{9.56}  & \textbf{25.51} \\
			\bottomrule
		\end{tabular}
	\end{center}
	\caption{\label{duc-table} \textsc{Rouge} recall evaluation results on DUC 2004 test set. All these models are tested using beam search. Results with \otherpaper{} mark are taken from the corresponding papers. The superscript \significant{}  indicates that our \ourModelName{} model performs significantly better than it as given by the 95\% confidence interval in the official \textsc{Rouge} script.}
\end{table}

\paragraph{MSR-ATC}

We report the full length \textsc{Rouge} F1 score on the MSR-ATC test set in Table \ref{msr-table}.
To the best of our knowledge, this is the first work that reports \textsc{Rouge} metric scores on the MSR-ATC dataset.
Note that we only compare our model with ABS since the others are not publicly available.
Our \ourModelName{} achieves 10.63 \textsc{Rouge}-2 F1 and outperforms the s2s+att baseline by 1.02 points.

\begin{figure*}[h]
	\centering
	\includegraphics[width=0.9\textwidth]{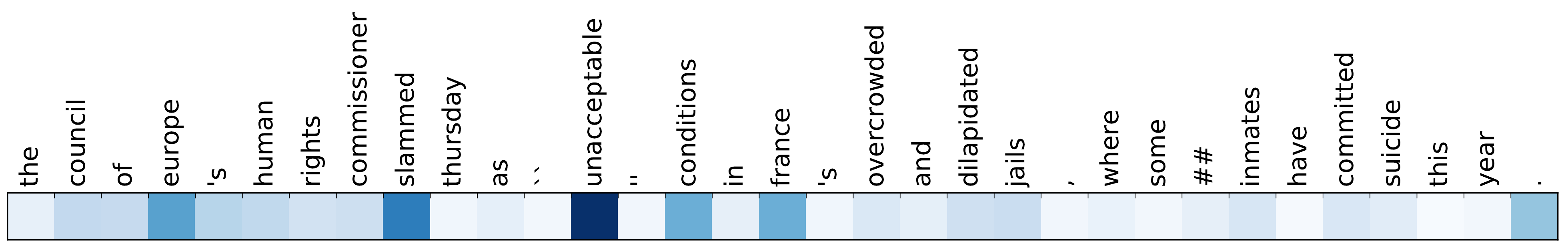}
	\caption{\label{fig:sal} First derivative heat map of the output with respect to the selective gate.
		The important words are selected in the input sentence, such as ``europe'', ``slammed'' and ``unacceptable''. The output summary of our system is ``council of europe slams french prison conditions'' and the true summary is ``council of europe again slams french prison conditions''.}
\end{figure*}

\begin{table}[htb]
	\begin{center}
		\begin{tabular}{lccc}
			\toprule
			\bf Models & \bf RG-1 & \bf RG-2 & \bf RG-L \\ 
			\midrule
			ABS (beam)  & 20.27\significant{} & 5.26\significant{} & 17.10\significant{} \\
			s2s+att (greedy)  & 15.15\significant{} & 4.48\significant{} & 13.62\significant{} \\
			s2s+att (beam)  & 22.65\significant{} & 9.61\significant{} & 21.39\significant{} \\
			\hline
			\ourModelName{} (greedy)  & 19.77 &  6.44 & 17.36 \\
			\ourModelName{} (beam)  & \textbf{25.75} &  \textbf{10.63} & \textbf{22.90} \\
			\bottomrule
		\end{tabular}
	\end{center}
	\caption{\label{msr-table} Full length \textsc{Rouge} F1 evaluation on MSR-ATC test set. Beam search are used in both the baselines and our method. The superscript \significant{}  indicates that our \ourModelName{} model performs significantly better than it as given by the 95\% confidence interval in the official \textsc{Rouge} script.}
\end{table}

\section{Discussion}
\label{sec:disc}

In this section, we first compare the performance of \ourModelName{} with the s2s+att baseline model to illustrate that the proposed method succeeds in selecting information and building tailored representation for abstractive sentence summarization.
We then analyze selective encoding by visualizing the heat map.

\paragraph{Effectiveness of Selective Encoding}
We further test the \ourModelName{} model with different sentence lengths on English Gigaword test sets, which are merged from the \citet{rush-chopra-weston:2015:EMNLP} test set and our internal test set.
The length of sentences in the test sets ranges from 10 to 80.
We group the sentences with an interval of 4 and get 18 different groups and we draw the first 14 groups.
We find that the performance curve of our \ourModelName{} model always appears to be on the top of that of s2s+att with a certain margin.
For the groups of 16, 20, 24, 32, 56 and 60, the \ourModelName{} model obtains big improvements compared to the s2s+att model.
Overall, these improvements on all groups indicate that the selective encoding method benefits the abstractive sentence summarization task.

\begin{figure}[htbp]
	\centering
	\includegraphics[width=0.5\textwidth]{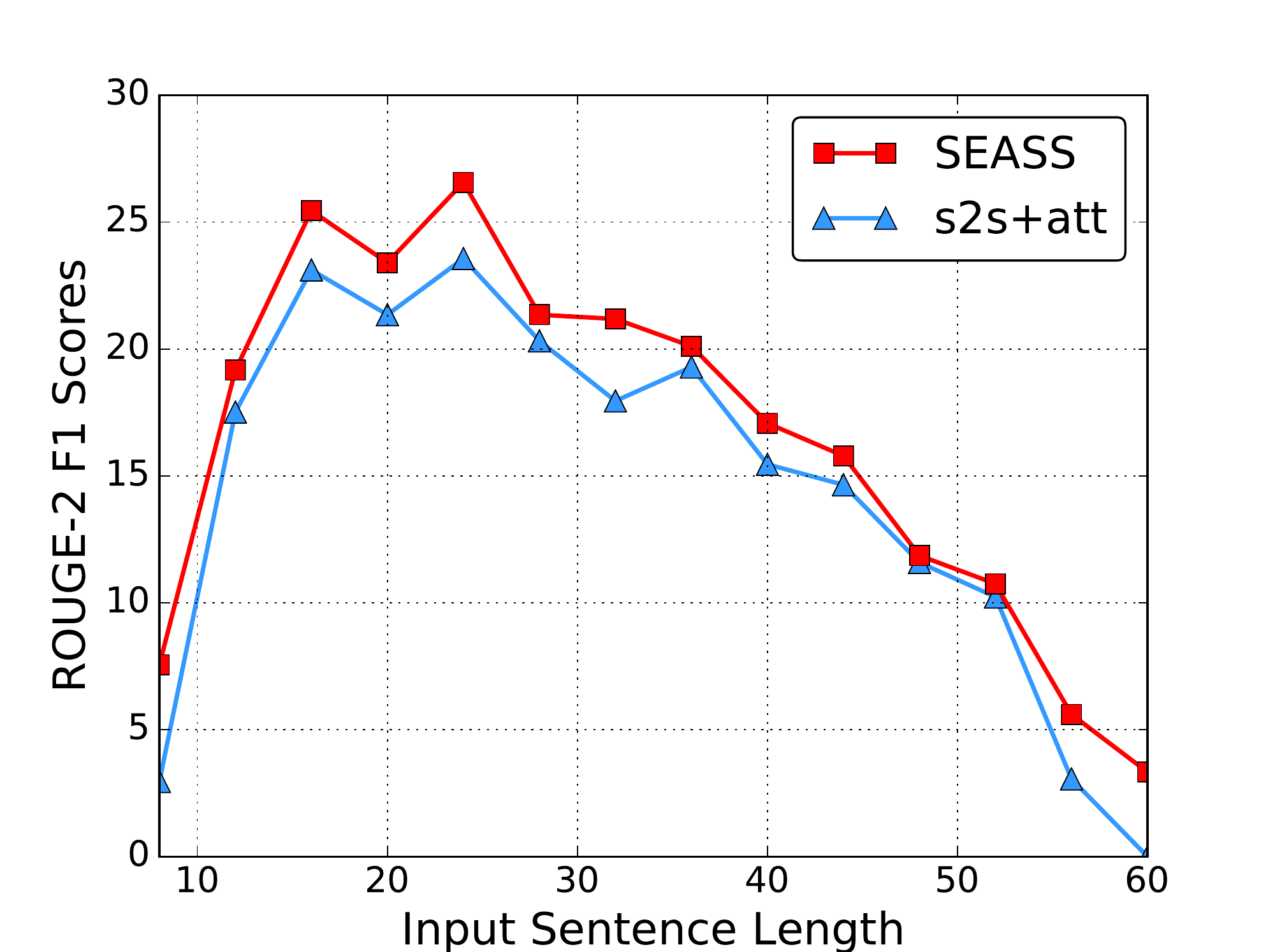}
	\caption{\label{fig:gigaRougeLen} \textsc{Rouge}-2 F1 score on different groups of input sentences in terms of their length for s2s+att baseline and our \ourModelName{} model on English Gigaword test sets.}
\end{figure}

\paragraph{Saliency Heat Map of Selective Gate}
Since the output of the selective gate network is a high dimensional vector, it is hard to visualize all the gate values.
We use the method in \citet{jiweiliNaacl} to visualize the contribution of the selective gate to the final output, which can be approximated by the first derivative.
Given sentence words $ x $ with associated output summary $ y $, the trained model associates the pair $ (x, y) $ with a score $ S_{y}(x) $.
The goal is to decide which gate $ g $ associated with a specific word makes the most significant contribution to $ S_{y}(x) $.
We approximate the $ S_{y}(g) $ by computing the first-order Taylor expansion since the score $ S_{y}(x) $ is a highly non-linear function in the deep neural network models:
\begin{equation}
S_{y}(g) \approx w(g)^{T} g + b
\end{equation}
where $ w(g) $ is first the derivative of $ S_{y} $ with respect to the gate $ g $:
\begin{equation}
w(g) = \frac{\partial(S_{y})}{\partial g} |_{g}
\end{equation}
We then draw the Euclidean norm of the first derivative of the output $ y $ with respect to the selective gate $ g $ associated with each input words.

Figure \ref{fig:sal} shows an example of the first derivative heat map, in which most of the important words are selected by the selective gate such as ``europe'', ``slammed'', ``unacceptable'', ``conditions'', and ``france''.
We can observe that the selective gate determines the importance of each word before decoder, which releases the burden of it by providing tailored sentence encoding.

\section{Conclusion}
This paper proposes a selective encoding model which extends the sequence-to-sequence model for abstractive sentence summarization task.
The selective mechanism mimics one of the human summarizers' behaviors, selecting important information before writing down the summary.
With the proposed selective mechanism, we build an end-to-end neural network summarization model which consists of three phases: encoding, selection, and decoding.
Experimental results show that the selective encoding model greatly improves the performance with respect to the state-of-the-art methods on English Gigaword, DUC 2004 and MSR-ATC test sets.

\section*{Acknowledgments}

We thank Chuanqi Tan, Junwei Bao, Shuangzhi Wu and the anonymous reviewers for their helpful comments.
We also thank Alexander M. Rush for providing the dataset for comparison and helpful discussions.

\bibliography{acl2017}

\begin{thebibliography}{}
\expandafter\ifx\csname natexlab\endcsname\relax\def\natexlab#1{#1}\fi

\bibitem[{Ayana et~al.(2016)Ayana, Shen, Liu, and
  Sun}]{DBLP:journals/corr/AyanaSLS16}
Ayana, Shiqi Shen, Zhiyuan Liu, and Maosong Sun. 2016.
\newblock Neural headline generation with minimum risk training.
\newblock {\em CoRR\/} abs/1604.01904.

\bibitem[{Bahdanau et~al.(2015)Bahdanau, Cho, and Bengio}]{bahdanau2014neural}
Dzmitry Bahdanau, Kyunghyun Cho, and Yoshua Bengio. 2015.
\newblock Neural machine translation by jointly learning to align and
  translate.
\newblock In {\em Proceedings of 3rd International Conference for Learning
  Representations\/}. San Diego.

\bibitem[{Banko et~al.(2000)Banko, Mittal, and Witbrock}]{banko2000headline}
Michele Banko, Vibhu~O Mittal, and Michael~J Witbrock. 2000.
\newblock Headline generation based on statistical translation.
\newblock In {\em Proceedings of the 38th Annual Meeting on Association for
  Computational Linguistics\/}. Association for Computational Linguistics,
  pages 318--325.

\bibitem[{Bengio et~al.(2003)Bengio, Ducharme, Vincent, and
  Jauvin}]{bengio2003neural}
Yoshua Bengio, R{\'e}jean Ducharme, Pascal Vincent, and Christian Jauvin. 2003.
\newblock A neural probabilistic language model.
\newblock {\em journal of machine learning research\/} 3(Feb):1137--1155.

\bibitem[{Cheng and Lapata(2016)}]{cheng-lapata:2016:P16-1}
Jianpeng Cheng and Mirella Lapata. 2016.
\newblock Neural summarization by extracting sentences and words.
\newblock In {\em Proceedings of the 54th Annual Meeting of the Association for
  Computational Linguistics (Volume 1: Long Papers)\/}. Association for
  Computational Linguistics, Berlin, Germany, pages 484--494.

\bibitem[{Cho et~al.(2014)Cho, van Merrienboer, Gulcehre, Bahdanau, Bougares,
  Schwenk, and Bengio}]{cho-EtAl:2014:EMNLP2014}
Kyunghyun Cho, Bart van Merrienboer, Caglar Gulcehre, Dzmitry Bahdanau, Fethi
  Bougares, Holger Schwenk, and Yoshua Bengio. 2014.
\newblock Learning phrase representations using rnn encoder--decoder for
  statistical machine translation.
\newblock In {\em Proceedings of the 2014 Conference on Empirical Methods in
  Natural Language Processing (EMNLP)\/}. Association for Computational
  Linguistics, Doha, Qatar, pages 1724--1734.

\bibitem[{Chopra et~al.(2016)Chopra, Auli, and
  Rush}]{chopra-auli-rush:2016:N16-1}
Sumit Chopra, Michael Auli, and Alexander~M. Rush. 2016.
\newblock Abstractive sentence summarization with attentive recurrent neural
  networks.
\newblock In {\em Proceedings of the 2016 Conference of the North American
  Chapter of the Association for Computational Linguistics: Human Language
  Technologies\/}. Association for Computational Linguistics, San Diego,
  California, pages 93--98.

\bibitem[{Dorr et~al.(2003)Dorr, Zajic, and Schwartz}]{dorr2003hedge}
Bonnie Dorr, David Zajic, and Richard Schwartz. 2003.
\newblock Hedge trimmer: A parse-and-trim approach to headline generation.
\newblock In {\em Proceedings of the HLT-NAACL 03 on Text summarization
  workshop-Volume 5\/}. Association for Computational Linguistics, pages 1--8.

\bibitem[{Glorot and Bengio(2010)}]{glorot2010understanding}
Xavier Glorot and Yoshua Bengio. 2010.
\newblock Understanding the difficulty of training deep feedforward neural
  networks.
\newblock In {\em Aistats\/}. volume~9, pages 249--256.

\bibitem[{Goodfellow et~al.(2013)Goodfellow, Warde-Farley, Mirza, Courville,
  and Bengio}]{goodfellow2013maxout}
Ian~J Goodfellow, David Warde-Farley, Mehdi Mirza, Aaron~C Courville, and
  Yoshua Bengio. 2013.
\newblock Maxout networks.
\newblock {\em ICML (3)\/} 28:1319--1327.

\bibitem[{Gu et~al.(2016)Gu, Lu, Li, and Li}]{gu-EtAl:2016:P16-1}
Jiatao Gu, Zhengdong Lu, Hang Li, and Victor~O.K. Li. 2016.
\newblock Incorporating copying mechanism in sequence-to-sequence learning.
\newblock In {\em Proceedings of the 54th Annual Meeting of the Association for
  Computational Linguistics (Volume 1: Long Papers)\/}. Association for
  Computational Linguistics, Berlin, Germany, pages 1631--1640.

\bibitem[{Gulcehre et~al.(2016)Gulcehre, Ahn, Nallapati, Zhou, and
  Bengio}]{gulcehre-EtAl:2016:P16-1}
Caglar Gulcehre, Sungjin Ahn, Ramesh Nallapati, Bowen Zhou, and Yoshua Bengio.
  2016.
\newblock Pointing the unknown words.
\newblock In {\em Proceedings of the 54th Annual Meeting of the Association for
  Computational Linguistics (Volume 1: Long Papers)\/}. Association for
  Computational Linguistics, Berlin, Germany, pages 140--149.

\bibitem[{Kingma and Ba(2015)}]{kingma2014adam}
Diederik Kingma and Jimmy Ba. 2015.
\newblock Adam: A method for stochastic optimization.
\newblock In {\em Proceedings of 3rd International Conference for Learning
  Representations\/}. San Diego.

\bibitem[{Knight and Marcu(2002)}]{knight2002summarization}
Kevin Knight and Daniel Marcu. 2002.
\newblock Summarization beyond sentence extraction: A probabilistic approach to
  sentence compression.
\newblock {\em Artificial Intelligence\/} 139(1):91--107.

\bibitem[{Li et~al.(2016)Li, Chen, Hovy, and Jurafsky}]{jiweiliNaacl}
Jiwei Li, Xinlei Chen, Eduard Hovy, and Dan Jurafsky. 2016.
\newblock Visualizing and understanding neural models in nlp.
\newblock In {\em Proceedings of the 2016 Conference of the North American
  Chapter of the Association for Computational Linguistics: Human Language
  Technologies\/}. Association for Computational Linguistics, San Diego,
  California, pages 681--691.

\bibitem[{Lin(2004)}]{lin2004rouge}
Chin-Yew Lin. 2004.
\newblock Rouge: A package for automatic evaluation of summaries.
\newblock In {\em Text summarization branches out: Proceedings of the ACL-04
  workshop\/}. Barcelona, Spain, volume~8.

\bibitem[{Luong et~al.(2015)Luong, Pham, and
  Manning}]{luong-pham-manning:2015:EMNLP}
Thang Luong, Hieu Pham, and Christopher~D. Manning. 2015.
\newblock Effective approaches to attention-based neural machine translation.
\newblock In {\em Proceedings of the 2015 Conference on Empirical Methods in
  Natural Language Processing\/}. Association for Computational Linguistics,
  Lisbon, Portugal, pages 1412--1421.

\bibitem[{Nallapati et~al.(2016)Nallapati, Zhou, glar Gul{\c{c}}ehre, and
  Xiang}]{nallapatiabstractive}
Ramesh Nallapati, Bowen Zhou, {\c{C}}a~glar Gul{\c{c}}ehre, and Bing Xiang.
  2016.
\newblock Abstractive text summarization using sequence-to-sequence rnns and
  beyond.
\newblock In {\em Proceedings of The 20th SIGNLL Conference on Computational
  Natural Language Learning\/}.

\bibitem[{Napoles et~al.(2012)Napoles, Gormley, and
  Van~Durme}]{Napoles:2012:AG:2391200.2391218}
Courtney Napoles, Matthew Gormley, and Benjamin Van~Durme. 2012.
\newblock Annotated gigaword.
\newblock In {\em Proceedings of the Joint Workshop on Automatic Knowledge Base
  Construction and Web-scale Knowledge Extraction\/}. Association for
  Computational Linguistics, Stroudsburg, PA, USA, AKBC-WEKEX '12, pages
  95--100.

\bibitem[{Over et~al.(2007)Over, Dang, and Harman}]{over2007duc}
Paul Over, Hoa Dang, and Donna Harman. 2007.
\newblock Duc in context.
\newblock {\em Information Processing \& Management\/} 43(6):1506--1520.

\bibitem[{Pascanu et~al.(2013)Pascanu, Mikolov, and
  Bengio}]{pascanu2013difficulty}
Razvan Pascanu, Tomas Mikolov, and Yoshua Bengio. 2013.
\newblock On the difficulty of training recurrent neural networks.
\newblock {\em ICML (3)\/} 28:1310--1318.

\bibitem[{Rush et~al.(2015)Rush, Chopra, and
  Weston}]{rush-chopra-weston:2015:EMNLP}
Alexander~M. Rush, Sumit Chopra, and Jason Weston. 2015.
\newblock A neural attention model for abstractive sentence summarization.
\newblock In {\em Proceedings of the 2015 Conference on Empirical Methods in
  Natural Language Processing\/}. Association for Computational Linguistics,
  Lisbon, Portugal, pages 379--389.

\bibitem[{Shen et~al.(2016)Shen, Cheng, He, He, Wu, Sun, and
  Liu}]{shen-EtAl:2016:P16-1}
Shiqi Shen, Yong Cheng, Zhongjun He, Wei He, Hua Wu, Maosong Sun, and Yang Liu.
  2016.
\newblock Minimum risk training for neural machine translation.
\newblock In {\em Proceedings of the 54th Annual Meeting of the Association for
  Computational Linguistics (Volume 1: Long Papers)\/}. Association for
  Computational Linguistics, Berlin, Germany, pages 1683--1692.

\bibitem[{Srivastava et~al.(2014)Srivastava, Hinton, Krizhevsky, Sutskever, and
  Salakhutdinov}]{srivastava2014dropout}
Nitish Srivastava, Geoffrey~E Hinton, Alex Krizhevsky, Ilya Sutskever, and
  Ruslan Salakhutdinov. 2014.
\newblock Dropout: a simple way to prevent neural networks from overfitting.
\newblock {\em Journal of Machine Learning Research\/} 15(1):1929--1958.

\bibitem[{Sutskever et~al.(2014)Sutskever, Vinyals, and
  Le}]{sutskever2014sequence}
Ilya Sutskever, Oriol Vinyals, and Quoc~V Le. 2014.
\newblock Sequence to sequence learning with neural networks.
\newblock In {\em Advances in neural information processing systems\/}. pages
  3104--3112.

\bibitem[{Toutanova et~al.(2016)Toutanova, Brockett, Tran, and
  Amershi}]{toutanova-EtAl:2016:EMNLP2016}
Kristina Toutanova, Chris Brockett, Ke~M. Tran, and Saleema Amershi. 2016.
\newblock A dataset and evaluation metrics for abstractive compression of
  sentences and short paragraphs.
\newblock In {\em Proceedings of the 2016 Conference on Empirical Methods in
  Natural Language Processing\/}. Association for Computational Linguistics,
  Austin, Texas, pages 340--350.

\bibitem[{Yu et~al.(2016)Yu, Buys, and
  Blunsom}]{yu-buys-blunsom:2016:EMNLP2016}
Lei Yu, Jan Buys, and Phil Blunsom. 2016.
\newblock Online segment to segment neural transduction.
\newblock In {\em Proceedings of the 2016 Conference on Empirical Methods in
  Natural Language Processing\/}. Association for Computational Linguistics,
  Austin, Texas, pages 1307--1316.

\bibitem[{Zajic et~al.(2007)Zajic, Dorr, Lin, and Schwartz}]{zajic2007multi}
David Zajic, Bonnie~J Dorr, Jimmy Lin, and Richard Schwartz. 2007.
\newblock Multi-candidate reduction: Sentence compression as a tool for
  document summarization tasks.
\newblock {\em Information Processing \& Management\/} 43(6):1549--1570.

\bibitem[{Zeng et~al.(2016)Zeng, Luo, Fidler, and Urtasun}]{zeng2016efficient}
Wenyuan Zeng, Wenjie Luo, Sanja Fidler, and Raquel Urtasun. 2016.
\newblock Efficient summarization with read-again and copy mechanism.
\newblock {\em arXiv preprint arXiv:1611.03382\/} .

\end{thebibliography}
\bibliographystyle{acl_natbib}

\end{document}